%% file: neurips_2024.tex
\documentclass{article}
\PassOptionsToPackage{numbers, compress}{natbib}

\usepackage{color, colortbl}
\usepackage{booktabs} 
\usepackage{multirow}
\usepackage{wrapfig}
\usepackage{appendix}
\usepackage{amsmath}
\usepackage[preprint,nonatbib]{neurips_2024}

\usepackage{natbib}  
\usepackage[utf8]{inputenc} 
\usepackage[T1]{fontenc}    
\usepackage{url}            
\usepackage{booktabs}       
\usepackage{amsfonts}       
\usepackage{nicefrac}       
\usepackage{microtype}      
\usepackage{xcolor}         
\usepackage{float}
\usepackage{graphicx}
\usepackage{hyperref}       
\usepackage{caption}
\usepackage{lipsum}  
\usepackage{float}
\usepackage{listings}
\usepackage{framed}
\title{NeuSpeech: Decode Neural signal as Speech}

\author{\parbox{13cm}
  {\centering
    {Yiqian Yang$^{1}$\thanks{These authors contributed equally to this work}  \ \ \ \ \ \ \  Yiqun Duan$^{2*}$ \ \ \ \ \ \ \ Qiang Zhang$^{1}$ \ \ \ \ \ \ \  Hyejeong Jo$^{3}$  \ \ \ \ \ \ \ Jinni Zhou$^{1}$ \ \ \ \ \ \ \ Won Hee Lee$^{3\dagger}$ \ \ \ \ \ \ \ Renjing Xu$^{1\dagger}$ \ \ \ \ \ \ \ Hui Xiong$^{1}$\thanks{These are corresponding authors} }\\ 
  }
}

\begin{document}
\maketitle
\footnotetext[1]{The Hong Kong University of Science and Technology (Guangzhou), People's Republic of China, \\Yiqian Yang and Hui Xiong are with AI Thrust, HKUST(GZ). \\Qiang Zhang and Renjing Xu are with MICS Thrust, HKUST(GZ), \\Jinni Zhou is with RBM Base, College of Future Technology,\\Email: yyang937@connect.hkust-gz.edu.cn, xionghui@hkust-gz.edu.cn, qzhang749@connect.hkust-gz.edu.cn, renjingxu@hkust-gz.edu.cn, eejinni@hkust-gz.edu.cn}
\footnotetext[2]{GrapheneX-UTS HAI Centre, Australia Artificial Intelligence Institute, University of Technology Sydney, Australia,\\Email:duanyiquncc@gmail.com}
\footnotetext[3]{Department of Software Convergence, Kyung Hee University, Republic of Korea,\\Email:girlsending0@khu.ac.kr, whlee@khu.ac.kr}
 

\begin{abstract}
Decoding language from brain dynamics is an important open direction in the realm of brain-computer interface (BCI), especially considering the rapid growth of large language models.  
Compared to invasive-based signals which require electrode implantation surgery, non-invasive neural signals (e.g. EEG, MEG) have attracted increasing attention considering their safety and generality. 
However, the exploration is not adequate in three aspects: 1) previous methods mainly focus on EEG but none of the previous works address this problem on MEG with better signal quality; 2) prior works have predominantly used ``teacher-forcing" during generative decoding, which is impractical; 3) prior works are mostly ``BART-based" not fully auto-regressive, which performs better in other sequence tasks.
In this paper, we explore the brain-to-text translation of MEG signals in a speech-decoding formation. 
Here we are the first to investigate a cross-attention-based ``whisper" model for generating text directly from MEG signals without teacher forcing.
Our model achieves impressive BLEU-1 scores of 60.30 and 52.89 without pretraining \& teacher-forcing on two major datasets (\textit{GWilliams} and \textit{Schoffelen}).
This paper conducts a comprehensive review to understand how speech decoding formation performs on the neural decoding tasks, including pretraining initialization, training \& evaluation set splitting, augmentation, and scaling law. Code is available at \href{https://github.com/NeuSpeech/NeuSpeech1}{https://github.com/NeuSpeech/NeuSpeech1}.

\end{abstract}

\section{Introduction}
\label{introduction}
Decoding language from brain activity is an area of neurotechnology that's rapidly evolving, offering significant benefits for improving communication and control for people with severe speech and motor impairments. Individuals with conditions like high-level spinal cord injuries or late-stage ALS often face significant communication challenges. Through noninvasive measurement and modeling of neural representations related to language production, we aim to enable digital communication based on thoughts alone. This advancement has the potential to transform the lives of people with disabilities, providing them with seamless and intuitive communication and device control capabilities.

Notably, using decoding speech on invasive signals achieved high-performance recognition accuracy for real-time translation~\cite{Metzger_2023_ecog_hubert_birnn_brain2speech_brain2text_avatar,Willett_2023_ecog_speech_neuroprosthesis_rnn_brain2speech2text}. However, the invasive recordings pose medical risks and issues with maintaining over long periods. Thus, for non-invasive brain recordings, Wang et al.~\cite{wang2022open_aaai_eeg2text} demonstrated open-vocabulary EEG-to-text translation by employing pre-trained language models on word-level EEG features. Duan et al.~\cite{duan2023dewave_brain2text} improved this approach by decoding from raw EEG waves without using time markers. However, limitations remain as evaluation relies on teacher forcing and the model cannot generate meaningful sentences without it. On MEG, previous works can merely decode limited classes from MEG signal.~\cite{dash2020decoding_meg_phrases,csaky2023interpretable_meg_many_class,ghazaryan2023trials_MEG_decode_written_text}. Recently, Defossez et al.\cite{D_fossez_2023_meg_eeg_clip_pretrain_meta_brain2speech} decoded speech perception of 3 second segments from MEG signals, and was capable of classifying sentences by matching MEG signal with speech. However, it cannot generate sentences from the MEG signal.

To address the issue of teacher forcing evaluation and generating text from MEG signal. In this paper, we present NeuSpeech, a pioneering framework that decodes neural as speech. NeuSpeech uses an encoder-decoder model (whisper~\cite{radford2023robust_whisper_model_originalpaper}) rather than BART-based models that are widely used in previous works~\cite{wang2022open_aaai_eeg2text,duan2023dewave_brain2text,xi2023unicorn} to directly translate raw MEG waves into text, without pretraining or transforming neural signals to discrete codes. This approach 1) lets the model efficiently learn text-related information from neural signals, and 2) keeps the general meaning of the whole sentence when generating long texts. 

Experiments employ non-invasive MEG signals from the large-scale public \textit{GWilliams}~\cite{Gwilliams_2023_dataset_meg_208sensors_27persons_56h} and \textit{Schoffelen}~\cite{Schoffelen_2019_dataset_meg_273sensor_96person_80h} datasets, which recorded MEG during speech listening tasks. Notably, NeuSpeech can generalize to all languages and equipment types. Performance is assessed using standard translation metrics~\cite{papineni2002bleu_orig,lin2004rouge_orig}. For raw MEG waves without event markers, NeuSpeech achieves 60.30 BLEU-1 and 55.26 Rouge-1 F-measure on \textit{GWilliams}, and 53.16 BLEU-1 on \textit{Schoffelen} without teacher forcing evaluation.
Decoding language from brain activity is an important and rapidly advancing area of neurotechnology. The ability to infer a person's intended words, phrases, or even full sentences directly from their brain signals has profound applications for improving communication and control capabilities for those suffering from severe speech and motor impairments. Conditions like high-level spinal cord injury or late-stage ALS can trap a person inside their own body, unable to fully express themselves or interact with the digital world around them. By non-invasively measuring and modeling the complex neural representations underlying language production in the brain, we aim to restore digital communication simply through thought. Beyond assisting those with disabilities, proficient decoding of full linguistic expressions from brain signals could also enable entirely new modes of seamless human-machine interaction. One could silently yet precisely command prosthetic devices, control software applications, browse the internet, or navigate virtual and augmented reality environments entirely using their brain. The potential applications seem nearly limitless and could vastly expand what is possible for both able-bodied individuals and those with severe impairments. It is an enormously challenging task from an artificial intelligence and neuroscience perspective but one with immense promise to transform lives if achieved.

Notably, using decoding speech on invasive signals achieved high-performance recognition accuracy for real-time translation~\cite{Metzger_2023_ecog_hubert_birnn_brain2speech_brain2text_avatar,Willett_2023_ecog_speech_neuroprosthesis_rnn_brain2speech2text}. However, the invasive recordings pose medical risks and issues with maintaining over long periods. Thus, for non-invasive brain recordings, Wang et al.~\cite{wang2022open_aaai_eeg2text} demonstrated open-vocabulary EEG-to-text translation by employing pre-trained language models on word-level EEG features. Duan et al.~\cite{duan2023dewave_brain2text} improved this approach by decoding from raw EEG waves without using time markers. However, limitations remain as evaluation relies on teacher forcing and the model cannot generate meaningful sentences without it. On MEG, previous works can merely decode limited classes from MEG signal.~\cite{dash2020decoding_meg_phrases,csaky2023interpretable_meg_many_class,ghazaryan2023trials_MEG_decode_written_text}. Recently, Defossez et al.\cite{D_fossez_2023_meg_eeg_clip_pretrain_meta_brain2speech} decoded speech perception of 3 second segments from MEG signals, and was capable of classifying sentences by matching MEG signal with speech. However, it cannot generate sentences from the MEG signal.

To address the issue of teacher forcing evaluation and generating text from MEG signal. In this paper, we present NeuSpeech, a pioneering framework that decodes neural as speech. NeuSpeech uses an encoder-decoder model (whisper~\cite{radford2023robust_whisper_model_originalpaper}) rather than BART-based models that are widely used in previous works~\cite{wang2022open_aaai_eeg2text,duan2023dewave_brain2text,xi2023unicorn} to directly translate raw MEG waves into text, without pretraining or transforming neural signals to discrete codes. This approach 1) lets the model efficiently learn text-related information from neural signals, and 2) keeps the general meaning of the whole sentence when generating long texts. 

Experiments employ non-invasive MEG signals from the large-scale public \textit{GWilliams}~\cite{Gwilliams_2023_dataset_meg_208sensors_27persons_56h} and \textit{Schoffelen}~\cite{Schoffelen_2019_dataset_meg_273sensor_96person_80h} datasets, which recorded MEG during speech listening tasks. Notably, NeuSpeech can generalize to all languages and equipment types. Performance is assessed using standard translation metrics~\cite{papineni2002bleu_orig,lin2004rouge_orig}. For raw MEG waves without event markers, NeuSpeech achieves 60.30 BLEU-1 and 55.26 Rouge-1 F-measure on \textit{GWilliams}, and 53.16 BLEU-1 on \textit{Schoffelen} without teacher forcing evaluation.

\section{Method}
NeuSpeech is a modification of whisper~\cite{radford2023robust_whisper_model_originalpaper} model, where the raw MEG signals are sent into the model and generate text. Whisper~\cite{radford2023robust_whisper_model_originalpaper} is the most powerful and noise-robust model in speech recognition which is trained on hundreds of thousands of audio data and can transcribe speech from most of the languages on earth. 

\subsection{Task Definition}

\label{sec:task_formulation}
Given a raw MEG waves $\mathcal{X}$, the aim is to decode the corresponding open-vocabulary text tokens $\mathcal{W}$. These MEG-Text pairs $\langle \mathcal{X}, \mathcal{W} \rangle$ are collected during hearing, as defined in Sec.~\ref{sec:data}. We decode neural signals as speech, and we only have one setting which is: Raw MEG Waves to Text Translation, where MEG waves $\mathcal{X}$ are directly sent into the whisper model to generate continuous words without teacher forcing, a more challenging but practical setting. NeuSpeech is the pioneering work in this task. Different from previous works\cite{duan2023dewave_brain2text,wang2022open_aaai_eeg2text}, we are the first to evaluate generated text without teacher forcing using MEG.

\subsection{Preprocessing}
We only implement basic pre-processing here, the MEG signals are notched at power line frequency and band-pass filtered within 1 Hz and 60 Hz. Then, it is resampled to 200Hz. After that, the signal is normalized using robust scaler mentioned in \cite{D_fossez_2023_meg_eeg_clip_pretrain_meta_brain2speech}, then clipped to $\pm 10$ and scaled into a value range of $(-1,1)$. 

\subsection{Model}
\label{sec.model}

\begin{figure}
    \centering
    \includegraphics[width=1\linewidth]{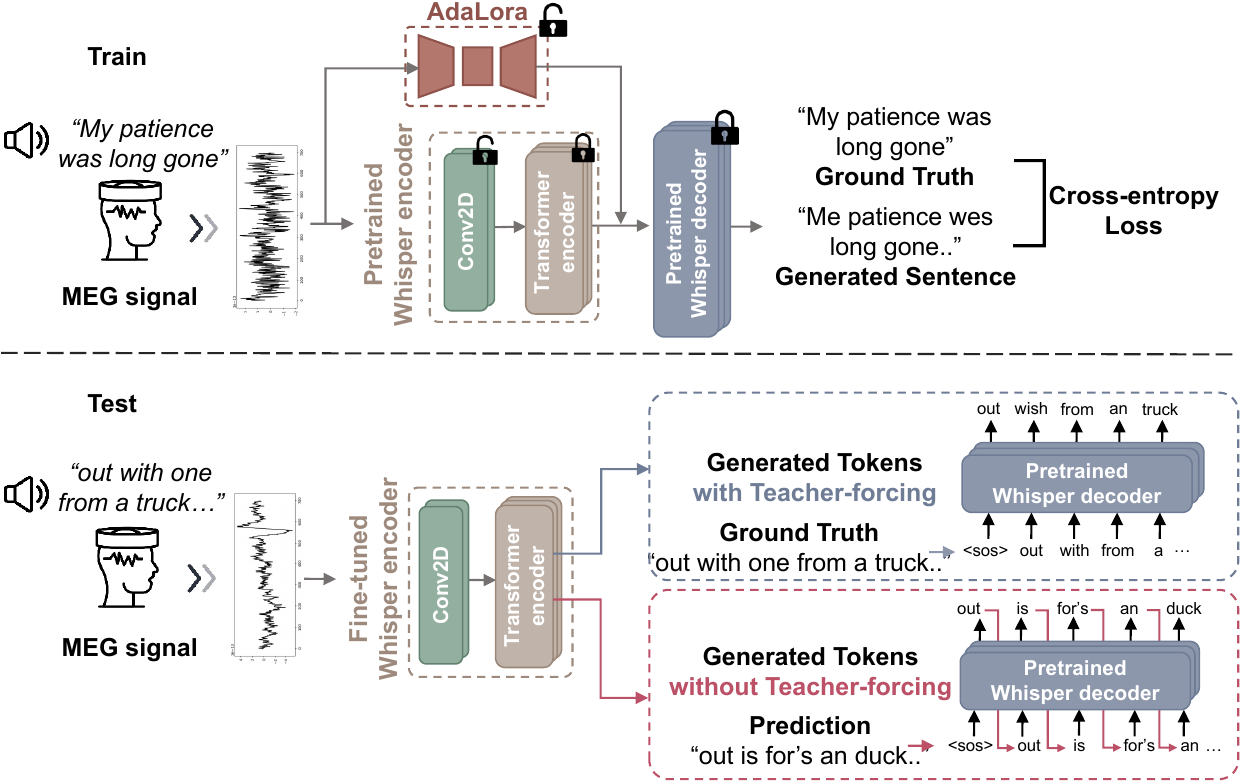}
    \caption{NeuSpeech overview. MEG signal is recorded while the subject is listening to speech. Our model is trained in an end-to-end manner, and only trains the AdaLora module applied on encoder and convolution layers using cross-entropy loss. In evaluation, we tested in both situations, w/ and w/o teacher forcing. In testing, teacher forcing means predict next token using previous ground truth tokens, rather than model generated tokens.}
    \label{fig:neuspeech_overview}
    \vspace{-25pt}
\end{figure}

Different from previous works~\cite{duan2023dewave_brain2text,Tang_2022_fmri_gpt_beam_search_brain2text} which are usually trained through multi-steps. Our model is trained end-to-end from scratch without bells and whistles. We freeze parameters in the decoder and only train the encoder. Since the original input channel of the whisper model is incompatible with the MEG channels, we replace the first convolution layer of the model with two convolution layers to fit in MEG channels and align time width respectively. For training data, we do not use any data augmentation. In terms of efficiency, we use Adalora~\cite{zhang2023adaptive_adalora} technique to tune our model, and only the encoder including convolution layers are in the Lora model.

\subsection{Evaluation}
\label{subsec:evaluation}

We evaluate transcribing performance on \textit{GWilliams} dataset~\cite{Gwilliams_2023_dataset_meg_208sensors_27persons_56h} and \textit{Schoffelen}~\cite{Schoffelen_2019_dataset_meg_273sensor_96person_80h} using NLP metrics, BLEU, ROUGE and WER, which are used in previous works~\cite{wang2022open_aaai_eeg2text,duan2023dewave_brain2text}. We add three baselines to give a rough understanding of up and down limits. 1) speech recognition, we use a pre-trained whisper base model to demonstrate the high performance on speech modality. 2) input noise, we input Gaussian noise into our model to check if our model can distinguish between notoriously noisy MEG and noise, and the well-trained model should not generate text with pure noise as input.

\section{Experiments}
\subsection{Dataset}\label{sec:data}
We utilize \textit{GWilliams} \cite{Gwilliams_2023_dataset_meg_208sensors_27persons_56h} and \textit{Schoffelen} \cite{Schoffelen_2019_dataset_meg_273sensor_96person_80h} datasets which are auditory MEG task for experiments. We only used auditory sentence data in Schoffelen. Both datasets have a detailed timetable for each sentence, and they are divided by sentence-recording pairs into train:validation:test 8:1:1, which results in 23339:2917:2918 and 8570:1112:1076 MEG-text pairs respectively. Unique words in GWilliams and Schoffelen is 2260 and 1861 respectively. We should note that the unique sentences are 661 and 360 for two datasets, which are very limited compared to speech datasets. More details at Supp.~\ref{supp.dataset}.



\subsection{Implementation Details }
\label{subsec:implementation details}
\textbf{Model modification.} We adopt whisper base model~\cite{radford2023robust_whisper_model_originalpaper} as our base model. To adapt the input size of the neural signal, we use two convolution1d layers, the kernel sizes are 3, and stride is 1,2 respectively to replace the first convolution layer in the whisper-base model of dimension 512. Detailed Python code at supp.\ref{supp.modification}.

\textbf{Training.} 
To implement eeg-to-text~\cite{wang2022open_aaai_eeg2text}, we use the original word timestamps provided by \textit{GWilliams}~\cite{Gwilliams_2023_dataset_meg_208sensors_27persons_56h} to segment MEG waves for each word feature. The training settings are the same as the origin code.

All models are trained on Nvidia H800(80GB) GPUs. We use a learning rate of 1e-3 and batch size of 64 for training 500 epochs and select the best model on evaluation loss. We use the AdamW as the optimizer for training all the models. We used early stopping on evaluation loss, if the evaluation loss does not decrease in 5 epochs, the training is stopped. It takes about 48 hours to train one model on single GPU.

\textbf{Evaluation.} For non-teacher forcing generation configuration, we use the beam search with a beam size of 5 and set the repetition penalty to 5, with no repeat n-gram size to 2, which is very important to avoid repeating.


\subsection{Metrics}

\input{text/table_score.tex}

As shown in Table~\ref{tb:scores}, we compare our results to input Gaussian noise to the NeuSpeech model and randomly select sentences from the test set and whisper base model speech transcription to give the understanding of up and down limits. We also compare previous method eeg-to-text~\cite{wang2022open_aaai_eeg2text}.

Based on the information provided in Tab.~\ref{tb:scores}, it is evident that our method outperforms randomly selected sentences, noise input, and eeg-to-text~\cite{wang2022open_aaai_eeg2text}. NeuSpeech has demonstrated the feasibility of decoding text from non-invasive brain recordings. Prior to NeuSpeech, many previous works~\cite{duan2023dewave_brain2text,wang2022open_aaai_eeg2text} in this field relied on teacher forcing to evaluate their models, Defossez et al.~\cite{D_fossez_2023_meg_eeg_clip_pretrain_meta_brain2speech} used classification accuracy. However, NeuSpeech introduces a more fair and real-world setting by employing end-to-end training, which directly ties neural signals to text. In comparison to previous works evaluated using teacher forcing, NeuSpeech has shown remarkable improvement, achieving BLEU-1, BLEU-2, BLEU-3, and BLEU-4 scores of 60.3, 55.26, 51.24, and 47.78, respectively, which is close to the performance of speech recognition.

To ensure a fair comparison, we also implement teacher forcing for both the eeg-to-text model~\cite{wang2022open_aaai_eeg2text} and NeuSpeech. It is evident that our BLEU-1 score of 80.20 is almost four times that of the eeg-to-text~\cite{wang2022open_aaai_eeg2text} model, highlighting the superiority of our approach. Furthermore, we have surpassed speech recognition methods with teacher forcing, emphasizing that our model can significantly improve performance by eliminating accumulated errors. It is worth noting that, at this stage, BLEU is the most suitable metric for evaluating the MEG-to-text system, as ROUGE and WER tend to yield high scores for noise input models.

\subsection{Generated Samples}
\input{text/table_targets.tex}
\input{text/table_targets_comparison}

In Tab.~\ref{tab:generation_results}, we present visual examples of text that have been generated from MEG signals. The training of \textit{Schoffelen}~\cite{Schoffelen_2019_dataset_meg_273sensor_96person_80h} is described in Sec.~\ref{subsubsec:abl:dataset} These examples showcase the capabilities of our model in generating meaningful text, despite the difficulties associated with translating thoughts and the limited prior research in this area. While our model may not yet achieve the same level of performance as nowadays speech recognition tasks, it demonstrates the ability to align keywords and construct sentences with similar structures. This is a promising indication of the potential of our approach in the field of brain-to-text translation.

The model's capability to match long sequences of words continuously is indeed impressive. As evidenced by the examples provided in Tab.~\ref{tab:generation_results}, such as the first sentence with 12 identical words, the model demonstrates the ability to accurately predict groups of words throughout a sentence. It is noteworthy that the model can leverage information from the MEG signals to make predictions, allowing it to generate correct groups of words even when there are occasional errors in the middle of a sentence. This highlights the model's capacity to capture and utilize contextual information from the brain signals, enabling it to generate meaningful and coherent text. Also, the model is good at transcribing Dutch which also shows the capability of generating continuous words.

We have provided Tab.~\ref{tab:generation_results_comparison} below, which compares the results of different approaches, namely NeuSpeech, NeuSpeech with teacher forcing, NeuSpeech with pre-training, and joint training. We observe that using NeuSpeech alone often result in semantically coherent sentences, even if not all the words were correct. Particularly in example (3), there was even a hint of expressive tone. However, when NeuSpeech was generating text with teacher forcing, although there were more correct word matches, there were often concatenated words, such as "onethe." The reason for this occurrence is not yet clear, but it significantly affects the semantic coherence of the sentences.

The performance of NeuSpeech improves to some extent when pre-training is employed. This improvement may be attributed, in part, to the model becoming more cautious and refraining from generating further output after a few errors. Among all the examples, only the sentences generated by the pre-trained model had word sequences without any incorrect words. This has both advantages and disadvantages. The advantage is that it improves the precision of the model and reduces the occurrence of incorrect words. However, the disadvantage is that once an error is predicted, the model becomes ineffective.

In contrast, the joint training model seems to tolerate longer sequences of word errors. For example, in (3), the model does not predict correct words for most of the sentence but accurately predicts the last eleven words, which is neglected by the higher-scoring pre-trained model. This indicates that, even for a model that reaches a certain performance level, we cannot solely rely on scoring metrics. Each model may have unique attributes that we can exploit. In addition, previous approaches for EEG-to-text translation seem to be less effective, even with the use of teacher forcing, they produced only a few words that match the true labels. This further highlights the superiority of our approach. In addition, the EEG-to-text model always predicts concepts like a long time, which is seen in supp.listing~\ref{supp.eeg_to_text_generation_wotf}.

In summary, the analysis of generated samples from different approaches reveals valuable insights, which emphasize the importance of considering both quantitative metrics and understanding the unique properties and strengths of different models. Furthermore, the comparison with previous EEG-to-text~\cite{wang2022open_aaai_eeg2text} methods underscores the advancements achieved by our approach.

\subsection{Ablation Study}\label{subsec:abl}
To demonstrate the effectiveness of our model and provide insight for large corpus training on this task. This study provided a comprehensive analysis of various factors influencing the performance of our model. 
The experimental results and theoretical insights obtained from these experiments not only validated the effectiveness of our model but also laid the groundwork for designing improved models, selecting appropriate training methods, devising better data augmentation techniques, and making informed model modifications.

%
\subsubsection{Different Dataset}\label{subsubsec:abl:dataset}\input{text/table_abl_dataset}

We evaluated our model in \textit{Schoffelen} dataset~\cite{Schoffelen_2019_dataset_meg_273sensor_96person_80h} to illustrate our effectiveness. As shown in Tab.~\ref{tab:abl_datasets}, our model has comparable performance on two datasets. This illustrates that our model is applicable under many scenarios, including English and Dutch, and can be adapted on different layouts.

\subsubsection{Pretrain}
\begin{wraptable}{r}{0.5\textwidth}
\centering
\caption{The performance of pretraining. The column Pretrain Dataset means trained on \textit{GWilliams}, \textit{Schoffelen}, and none. \textbf{Bold} for highest socres.}
\resizebox{0.95\linewidth}{!}{
\begin{tabular}{llllll}
\toprule
\multicolumn{1}{c}{\multirow{2}{*}{\begin{tabular}[c]{@{}c@{}}Pretrain \\ Dataset\end{tabular}}} & \multicolumn{1}{c}{\multirow{2}{*}{\begin{tabular}[c]{@{}c@{}}Finetune \\ Dataset\end{tabular}}} & \multicolumn{4}{c}{BLEU-N ($\%$) $\uparrow$} \\ \cmidrule{3-6} 
\multicolumn{1}{c}{}                          & \multicolumn{1}{c}{}                                                                             & N=1    & N=2    & N=3   & N=4   \\ \midrule
\textit{GWilliams}                                     & \textit{Schoffelen}                                                                                       & 48.09   & 42.23    & 38.5   & 35.60   \\
\textit{Schoffelen}                                    & \textit{GWilliams}                                                                                        & \textbf{63.68}    & \textbf{61.06}   & \textbf{59.26}   & \textbf{57.76}   \\
-                                             & \textit{Schoffelen}                                                                                        & 52.89  & 48.53  & 46.02 & 44.18\\
-                                             & \textit{GWilliams}                                                                                        & 60.30  & 55.26  & 51.24 & 47.78\\
\bottomrule
\end{tabular}
}
\label{tab:pretrain_effect}
\end{wraptable}

Here pretrain means training on one MEG dataset, and continuing training on another MEG dataset. Specifically, we trained the model on the \textit{Schoffelen} dataset~\cite{Schoffelen_2019_dataset_meg_273sensor_96person_80h} and fine-tuned on \textit{GWilliams}~\cite{Gwilliams_2023_dataset_meg_208sensors_27persons_56h}, and vice versa. As shown in Table \ref{tab:pretrain_effect}, pretraining provided benefits in some settings.

Interestingly, fine-tuning on \textit{GWilliams} after pretraining achieved higher performance than training from scratch, while pretraining on \textit{GWilliams} and fine-tuning on \textit{Schoffelen} yielded much lower scores. One potential reason for this asymmetry is that \textit{GWilliams} contains over twice as many samples as \textit{Schoffelen}. Thus, after pretraining on the larger \textit{GWilliams} set, the model is already highly optimized, making it difficult for the training loss to further descend during fine-tuning on the smaller \textit{Schoffelen} data.

In contrast, fine-tuning the model pre-trained on \textit{Schoffelen} Still allowed for improvements to be made on the larger \textit{GWilliams} distributions. This suggests pretraining can help in low-data regimes but may inhibit learning when moving to a dataset with a broader sample scope. Overall, these pretraining experiments provide insights into how architectural inductive biases interact with dataset characteristics to influence translation performance.

\subsubsection{Joint Training}
\begin{wraptable}{r}{0.5\textwidth}
\centering
\setlength{\tabcolsep}{3.9pt}
\caption{Performance of joint training on two datasets.}
\label{tab:joint training}
\resizebox{1\linewidth}{!}{
\begin{tabular}{@{}llllllll@{}}
\toprule
\multicolumn{1}{c}{\multirow{2}{*}{Dataset}} & \multicolumn{4}{c}{BLEU-N (\%)} & \multicolumn{3}{l}{ROUGE-1 (\%)} \\ \cmidrule(l){2-8} 
\multicolumn{1}{c}{}                         & N=1    & N=2    & N=3   & N=4   & F         & P         & R        \\ \cmidrule(r){1-8}
\textit{GWilliams}                                    & 55.13  & 52.91  & 51.28 & 49.89 & 57.33     & 64.22     & 57.17    \\
\textit{Schoffelen}                                   & 45.12  & 39.00  & 35.48 & 33.03 & 45.15     & 46.47     & 45.49    \\ \bottomrule
\end{tabular}
}
\end{wraptable}
Joint training means training with multiple datasets at one time. We want to find out if the model can learn from multiple datasets together which is a key to enable large-pretraining of models. It is important to note that the \textit{GWilliams} dataset comprises only 208 channels, thus we used the simplest method of applying padding on the channel dimensions to match the \textit{Schoffelen}'s channel dimension of 273 to see if this simple method works out. 


We got 55.13 and 45.12 BLEU-1 scores on \textit{GWilliams} and \textit{Schoffelen} respectively, which is 5-7 points lower compared to training solely from scratch. The results demonstrate that, although the performance was slightly affected, the simple data padding method was able to effectively fuse the two datasets. 

However, we acknowledge that the current method we employed is rudimentary, and future work will involve exploring more advanced techniques for dataset integration. This includes incorporating physical information from the sensors and employing more sophisticated approaches to merge the datasets. By doing so, we can further enhance the performance and effectiveness of joint training in multi-language and multi-layout neural transcribing systems.

\subsubsection{Scaling Law}
\label{subsubsec:scalinglaw}

\begin{wrapfigure}{r}{0.5\textwidth}
    
    \centering
    \includegraphics[width=\linewidth]{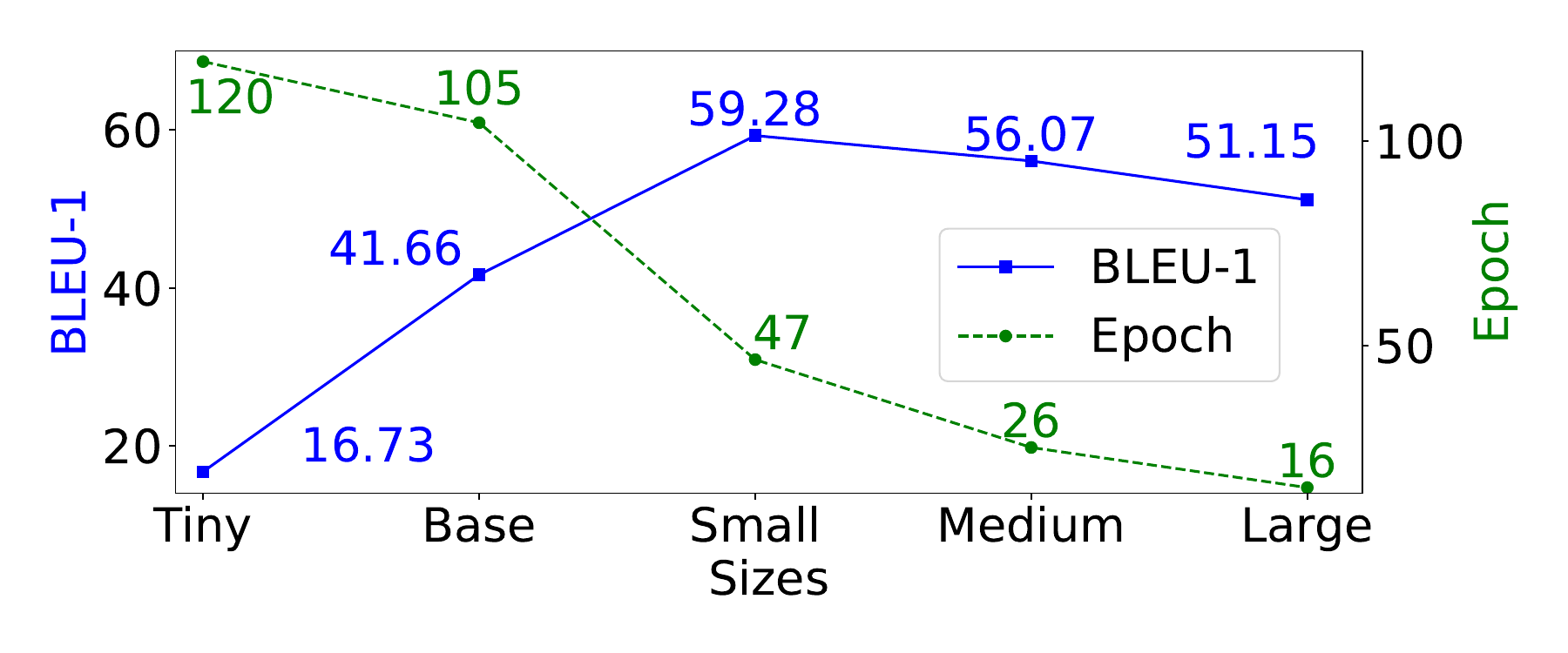}
    \caption{Performance on different model sizes. Black numbers represent the BLEU-1 score, green numbers are effective epochs after which the evaluation loss does not descend. Note that each experiment runs 120 epochs.}
    \label{fig:performance_with_sizes}
\end{wrapfigure}
To investigate the presence of a scaling law phenomenon in our model, where larger model sizes correspond to improved performance, we conduct an experiment using the Whisper series models at various sizes: tiny, base, small, medium, and large. We limit the GPU resources for training to a single H800 card with 80 GiB of memory, which results in batch size values of 128, 64, 16, 8, and 4, from smallest to largest model.


This experiment utilizes the \textit{Schoffelen} dataset and sets a maximum training epoch of 120. The best-performing checkpoint is selected for each model size. The results, depicted in Fig.~\ref{fig:performance_with_sizes}, indicate that when faced with limited data, increasing the model size leads to a consistent improvement in model performance until a certain size is reached.


Additionally, we observe that the number of epochs required to achieve convergence decreases by approximately half as the model size increases. 
By carefully balancing the bias-variance trade-off for each specific task and dataset, researchers can fine-tune the model to generalize optimally within the limitations of the available data, thereby maximizing decoding quality given the computational resources at hand.

\subsubsection{Data augmentation}

There are a lot of research works that used data augmentation in EEG/MEG signal, Bai et al.~\cite{bai2023dreamdiffusion} used time-wise masking. Now, we explore three kinds of augmentation, which are masking, noise injecting, and shifting, and compare them with baseline using Whisper-Base in Sec.~\ref{subsubsec:scalinglaw} which does not have data augmentation. We set three types of masking, which are time-wise, channel-wise, and block-wise. The mask unit is [channel,40],[1, time],[1,40], for the three masking methods respectively. We set two signal-to-noise ratios (SNR), 0dB and 15dB, of Gaussian noise with a probability of 0.5 and 1, which compose 4 settings. We also set 0.5 and 1 probability of data shifting to see if moving the neural data uniformly in the range of max input duration, along with the timestamps can help the model reduce over-fitting.

\begin{wrapfigure}{r}{0.5\textwidth}
    \centering
    \includegraphics[width=1\linewidth]{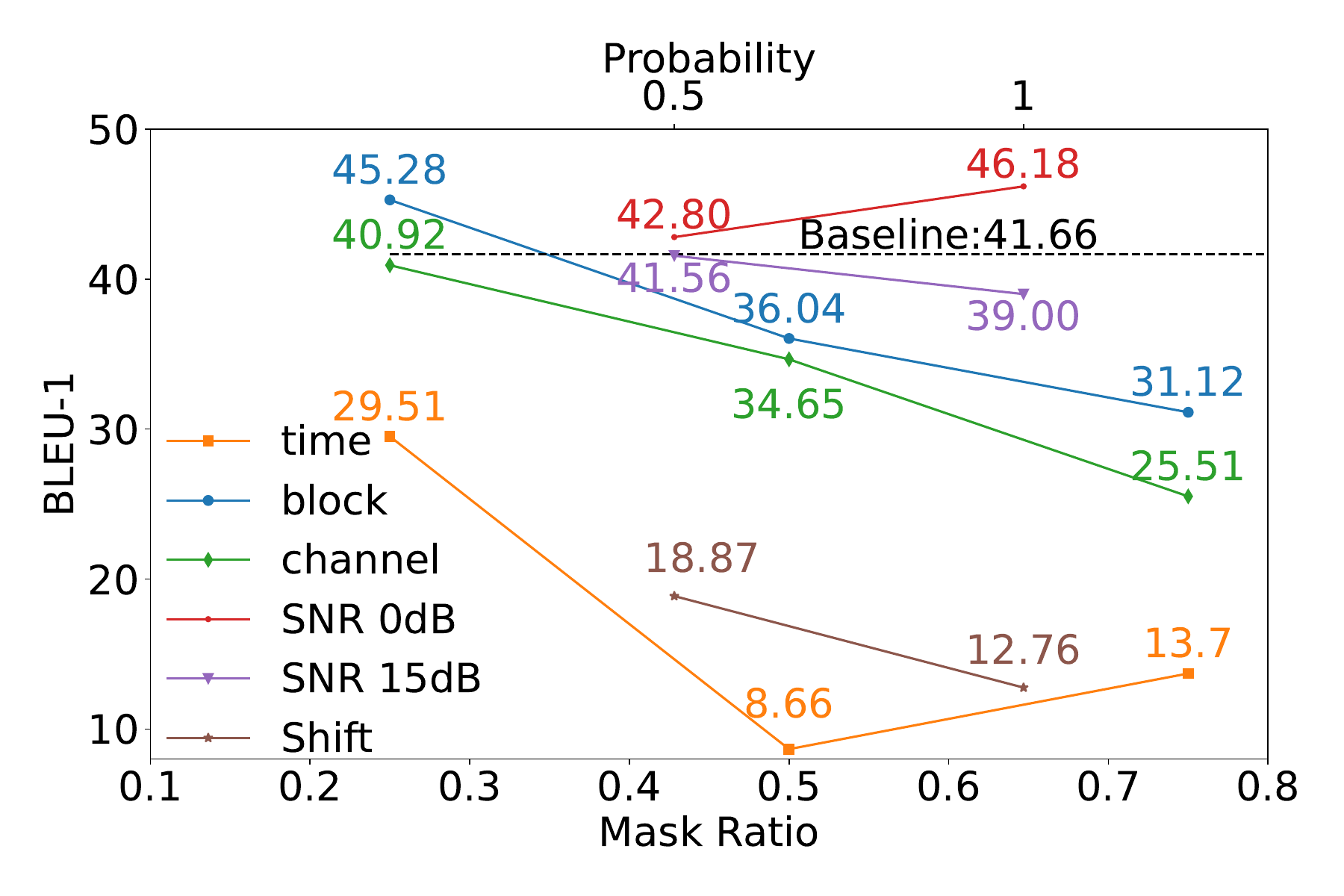}
    \caption{Performance changes with different data augmentations. The probability means the likelihood of adding augmentation to each data segment.}
    \label{fig:performance_with_da}
\end{wrapfigure}

Fig.~\ref{fig:performance_with_da} shows that block masking can improve performance a lot when the mask ratio is relatively low. However, when the masking ratio is bigger, the performance seems to decrease quickly. It can also be noticed that time masking is much worse than the other two mask types to our surprise. It may indicate that the redundancy of the MEG signal is much lower at this sampling rate compared to images which usually use a mask ratio of 75\%~\cite{he2022masked_mae}, when the mask ratio is too big to lose essential information, it can be hard for the model to recover information back correctly. 
Shifting gains no improvements under our experimental setting, oppositely it degrades model performance significantly. 
To our surprise, adding strong noise which is 0dB can enhance the model's performance greatly. But for weak noise of SNR 15dB, there is no benefit in our experimental setting. It reminds us that the MEG signal is very noisy, thus using strong noise may let the model better eliminate noise. Therefore, we recommend utilizing block masking with a small mask ratio and low signal-to-noise ratio (SNR) noise for data augmentation as they have demonstrated their effectiveness in enhancing the performance of the models.


    


\subsubsection{Data ratio and Fine-tune layers}

\begin{wrapfigure}{r}{0.5\textwidth}
    \vspace{-25pt}
    \centering
    \includegraphics[width=\linewidth]{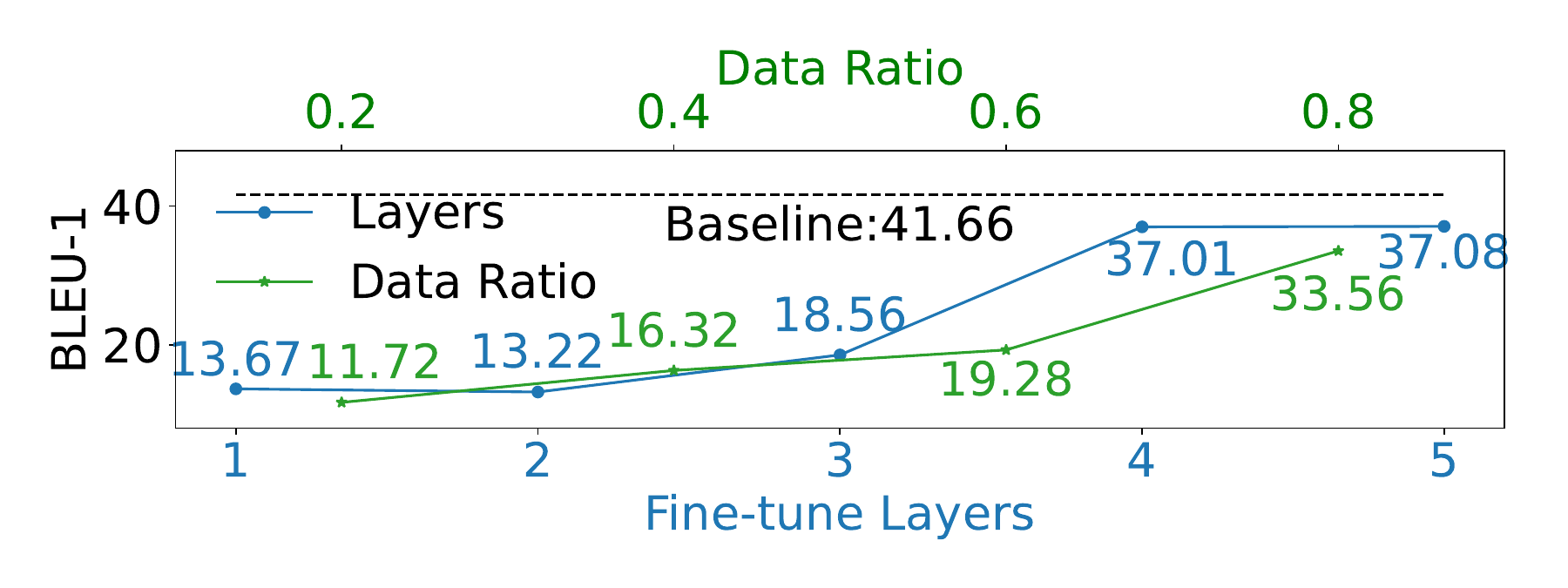}
    \caption{Fine-tune layers and data ratio. Blue line for different fine-tune layers, green line for different training data ratios. Note the whisper-base model only has 6 layers.}
    \label{fig:layers_and_data_ratio}
\end{wrapfigure}

We explore the ratio and data used in training, locking validation set and test set to ensure fairness. In Fig.~\ref{fig:layers_and_data_ratio}, we can see that performance rises as data is added as expected. Higher performance may be achieved if more data is provided since the curve seems far behind reaching the plateau. 

It can be observed that the more layers we fine-tune, the better the model will be. Note, that it is only on this base model, and may not be applicable to all. 


\section{Limitations}
\label{sec.limitation}
While NeuSpeech shows promising results on benchmark MEG-to-text tasks, there remain opportunities to strengthen the scientific validity and real-world applicability of the framework. Future work could focus on incorporating less structured brain signals to better validate the translatability of direct human thought, including physical sensor position data for improved accuracy, and exploring augmented datasets to address data scarcity issues that currently limit the model's ability to generalize and generate completely correct novel sentences. Addressing these limitations through collecting additional training data and enriching the inputs could help advance this emerging field by propelling NeuSpeech towards decoding thought from brain signals in varied real-world scenarios.
\section{Conclusion}

This paper explores MEG-to-text using existing speech recognition training, introducing the concept of decoding neural signals as speech. This approach brings high performance on two datasets on standard metrics and stretches to decode MEG signals without using teacher forcing. We also demonstrate the feasibility of pre-training or jointly training large neural speech models across layouts and languages. We have also explored the effects of augmentation and so on. Despite these advances, the results of hold-out scenarios are relatively low due to the limited number of sentences and the noisy nature of the MEG signal. In our future work, we will solve the problem of limited training sets and also find more mechanisms to fuse multi-modalities, and multi-layers with physical information embeddings.
\clearpage

\medskip

{
\small
\bibliography{IEEEabrv, references}
}
\newpage
\appendix
\section*{\Large \centering Supplementary Material for \\NeuSpeech: Decode Neural signal as Speech}

\section{Dataset}
\label{supp.dataset}

The Gwilliams~\cite{Gwilliams_2023_dataset_meg_208sensors_27persons_56h} dataset is created by Laura Gwilliams and colleagues. Below is the dataset detail quoted from their paper. \begin{quote}
\textbf{Participants} 
Twenty-seven English-speaking adults were recruited from the subject pool of NYU Abu Dhabi (15 females; age: M=24.8, 
SD=6.4). All participants provided written informed consent and were compensated for their time. Participants reported 
having normal hearing and no history of neurological disorders. All but one participants (S20) were native English speakers. 
All but five participants (S3, S11, S15, S19, S20) performed two identical one-hour-long sessions. These two recording sessions 
were separated by at least one day and at most two months depending on the availability of the experimenters and of the 
participants. The study was approved by the Institution Review Board (IRB) ethics committee of New York University Abu 
Dhabi. 
 
\textbf{Procedure} 
Within each ~1 h recording session, participants were recorded with a 208 axial-gradiometer MEG scanner built by the 
Kanazawa Institute of Technology (KIT), and sampled at 1,000 Hz, and online band-pass filtered between 0.01 and 200 Hz 
while they listened to four distinct stories through binaural tube earphones (Aero Technologies), at a mean level of 70 dB sound 
pressure level. 
Before the experiment, participants were exposed to 20 sec of each of the distinct speaker voices used in the study to (i) 
clarify the structure of the session and (ii) familiarize the participants with these voices. 
The order in which the four stories were presented was assigned pseudo-randomly, thanks to a "Latin-square design" across 
participants. This participant-specific order was used for both recording sessions. 
To ensure that the participants were attentive to the stories, they answered, every ~3 min and with a button press, a two- 
alternative forced-choice question relative to the story content (e.g. ‘What precious material had Chuck found? Diamonds or 
Gold’). Participants performed this task with an average accuracy of 98\%, confirming their engagement with and comprehension 
of the stories. 
Participants who did not already have a T1-weighted anatomical scan usable for the present study were scanned in a 3T 
Magnetic-Resonance-Imaging (MRI) scanner after the MEG recording to avoid magnetic artefacts. Six participants did not 
return for their T1 scan. 
Before each MEG session, the head shape of each participant was digitized with a hand-held FastSCAN laser scanner 
(Polhemus), and co-registered with five head-position coils. The positions of these coils with regard to the MEG sensors 
were collected before and after each recording and stored in the ’marker’ file, following the KIT’s system. The experimenter 
continuously monitored head position during the acquisition to ensure that the participants did not move. 

\textbf{Stimuli} 
Four English fictional stories were selected from the Open American National Corpus – a manually annotated corpus 
distributed without license or other restrictions1: 

• ‘Cable spool boy‘: a 1,948-word story narrating two young brothers playing in the woods 

• ‘LW1‘: a 861-word story narrating an alien spaceship trying to find its way home 

• ‘Black willow‘: a 4,652-word story narrating the difficulties an author encounters during writing. 

• ‘Easy money‘: a 3,541-word fiction narrating two friends using a magical trick to make money. 

An audio track corresponding to each of these stories was synthesized using Mac OS Mojave’s (c) text-to-speech. To help 
decorrelate language from acoustic representations, we varied both voices and speech rate every 5-20 sentences. Specifically, 
we used three distinct synthetic voices, namely ‘Ava‘, ‘Samantha‘ and ‘Allison‘ speaking between 145 and 205 words per 
minute. Additionally, we varied the silence between sentences between 0 and 1,000 ms. 
Each story was divided into around 5 min sound files. In between these sounds – approximately every 30 s – we played a random 
word list generated from the unique content words (nouns, proper nouns, verbs, adverbs and adjectives) of the preceding 5 min 
segment presented in random order. In addition, a very small fraction (<1\%) of non-words were introduced in natural sentences. 
Hereafter, and following the BIDS labeling, each "task" corresponds to the concatenation of these sentences and word 
lists. Each subject listened to the exact same set of four tasks, only in a different block order.
\end{quote}

The Schoffelen dataset is a multimodal neuroimaging dataset, below is the detailed description from their paper~\cite{Schoffelen_2019_dataset_meg_273sensor_96person_80h}. Note we only used auditory sentence in our experiments.
\begin{quote}
\textbf{Subjects}
A total of 204 native speakers of Dutch (100 males) with a mean age of 22 years (range: 18 to 33 years) were included in the study. In the informed consent procedure, they explicitly consented for the anonymized collected data to be used for research purposes by other researchers. The subjects took part in both the fMRI and MEG part of the study, in a counterbalanced fashion. Each subject performed the task in either the visual or the auditory modality. All subjects were right-handed, had normal or corrected-to-normal vision, and reported no history of neurological, developmental or language deficits. The study was approved by the local ethics committee (CMO – the local “Committee on Research Involving Human Subjects” in the Arnhem-Nijmegen region) and followed guidelines of the Helsinki declaration.

\textbf{Stimulus material}
The total stimulus set consisted of 360 sentences in Dutch, and their scrambled word list counterparts. The sentences consisted of two types: 180 of the sentences contained a relative clause (RC+), to create a more difficult syntactic structure. The other 180 sentences consisted of a main clause and a simple subordinate clause (RC-), to create an easier structure. The word lists were created by scrambling the words from the sentences such that three or more consecutive words did not form a coherent fragment. For an example of the sentences see Table 1. All sentences varied between 9 and 15 words in length.

Each subject was presented with a subset of 180 sentence stimuli, and 180 word list stimuli, where we ensured that, for a given sentence, they were not exposed to the corresponding word list. During the task MEG part of the experiment, the subjects were presented with 120 sentences and 120 word lists, and during the fMRI part of the experiment they were presented with 60 sentences and 60 word lists. Across subjects, all stimuli were presented the same number of times in the sentence and in the word list condition.

Each sentence and corresponding word list contained a noun that was at the same ordinal position, which varied between the third and thirteenth word position (which is denoted here as the ‘target word’). This allows for psychometrically controlled comparisons between the conditions using these isolated nouns. These words were matched in terms of lexical frequency and word length. To control for similarity in context, the word preceding the target did not differ in word length by more than two letters between a sentence and its word list counterpart. The word frequency was based on the lemma frequency according to the SUBTLEX-NL database of Dutch word frequencies21, and was on average 27,6/million (SD = 62,1/million). All stimulus material is shared along with the neuroimaging data.

The auditory versions of the stimuli were recorded by a female native Dutch speaker. The word lists were pronounced with a neutral prosody and a clear pause in between each word. The audio files were recorded in stereo at 44100 Hz. During the post processing the audio files were low-pass filtered at 8500 Hz and normalized such that all audio files had the same peak amplitude, and same peak intensity. In the word list condition, each word was separated by 300 ms of total silence. The transition from silence to speech was ramped at the onset (rise time of 10 ms) and offset (fall time of 10 ms) of single words in the word list condition, and for sentence onset. The onset of the first and target word vocalizations were determined manually for each audio file, using the Praat software (\text{http://www.praat.org, RRID:SCR\_016564}).

\textbf{Experimental Design and Procedure}
The total stimulus set was divided into two sets of 180, and each of these were subsequently divided into three subsets. Each subject was presented with 2/3 of the stimuli set in the MEG (120 trials of each condition) and 1/3 in the fMRI (60 trials). Across subjects, each subset was presented as many times in MEG as in fMRI. Different subjects that were presented with the same subset had the stimuli presented in a different (randomized) order. Within an experimental session, the stimuli were presented in blocks, alternating between sentence blocks (containing 5 sentences) and word list blocks (containing 5 word lists), for a total of 24 (MEG) or 12 (fMRI) blocks. The starting block type (either sentences or word list) was randomized across subjects.

In order to check for compliance, 20\% of the trials were followed by a ‘Yes’/‘No’ question about the content of the previous sentence/word list. Half of the questions on the sentences addressed the content of the sentence (e.g. Did grandma give a cookie to the girl?) whereas the other half, and all of the questions about the word lists, addressed one of the main content words (e.g. Was the word ‘grandma’ mentioned?). A substantial part of the questions on complex relative clause (RC+) sentences concerned content from the relative clause. Subjects answered the question by pressing a button for ‘Yes’/‘No’ with their left index and middle finger, respectively.

At the start of each block there was a 1500 ms presentation of the block type: zinnen (sentences) or woorden (words). In sentences, the first word began with a capital letter, and the last word end with a full stop. The inter-trial interval was jittered between 3200–4200 ms. During this period, an empty screen was presented, followed by a fixation cross.

Stimuli were presented using the Presentation software (Version 16.0, Neurobehavioral Systems, Inc).

Prior to the task, subjects read a written instruction of the task and were allowed to ask questions for clarification. Furthermore, the experimenter emphasized that the sentences and word lists should be attended carefully, and discouraged attempts to integrate the words in the word list condition. Finally, to familiarize the subjects with the task, they did a practice task with stimuli separate from the actual study task. Figure 1 shows a schematic overview of the study procedure.

\textbf{Auditory language task}
In the auditory task the stimuli were presented via plastic tubes and ear pieces to both ears. Before the experiment, the hearing threshold was determined individually and the stimuli were then presented at an intensity of 50 dB above the hearing threshold. In fMRI, the hearing threshold was determined on top of the EPI-sequence noise, to verify that all stimuli were clearly audible.

\end{quote}

\newpage
We arranged our preprocessed data into .jsonl files, here is an example.
\begin{lstlisting}
{"speech": {"path": "/hpc2hdd/home/yyang937/datasets/gwilliams2023/preprocess5/sub-21/ses-0/meg/sub-21_ses-0_task-3_senid_11_audio.wav", "sr": 16000}, "eeg": {"path": "/hpc2hdd/home/yyang937/datasets/gwilliams2023/preprocess5/sub-21/ses-0/meg/sub-21_ses-0_task-3_senid_11_meg.npy", "sr": 200}, "duration": 3.700000000000003, "language": "English", "sentence": "Although the concept was a simple one Allan thought it had potential", "sentences": [{"text": "Although the concept was a simple one Allan thought it had potential", "start": 0.0, "end": 3.700000000000003, "duration": 3.700000000000003, "words": [{"word": "Although", "start": 24.653999999999996, "end": 24.964}, {"word": "the", "start": 24.964, "end": 25.073999999999998}, {"word": "concept", "start": 25.073999999999998, "end": 25.593999999999994}, {"word": "was", "start": 25.593999999999994, "end": 25.733999999999995}, {"word": "a", "start": 25.733999999999995, "end": 25.803999999999988}, {"word": "simple", "start": 25.804000000000002, "end": 26.12400000000001}, {"word": "one", "start": 26.423999999999992, "end": 26.774}, {"word": "Allan", "start": 26.933999999999997, "end": 27.233999999999995}, {"word": "thought", "start": 27.233999999999995, "end": 27.453999999999994}, {"word": "it", "start": 27.453999999999994, "end": 27.533999999999992}, {"word": "had", "start": 27.543999999999997, "end": 27.72399999999999}, {"word": "potential", "start": 27.724000000000004, "end": 28.354000000000013}]}], "subj": 21, "story": "The_Black_Willow", "story_id": 3.0, "seq_id": 11.0, "sound_id": 0.0, "speech_rate": 205.0, "voice": "Samantha", "start": 92.824, "end": 96.52400000000002, "audio_start": 68.17, "audio_end": 71.87}
\end{lstlisting}

\section{Splits}
\label{supp.splits}
\input{text/table_split_datasets.tex}
We conducted further experiments using the \textit{GWilliams}~\cite{Gwilliams_2023_dataset_meg_208sensors_27persons_56h} dataset to evaluate NeuSpeech's generalizability.  
To test the model's ability to generalize across recording sessions, subjects, and unique sentences, we developed three additional training/validation splits beyond the standard 8:1 split. Specifically, we withheld: 1) all data from the second recording session, 2) all instances of the story ``lw1", and 3) a random 10\% sample of unique sentences. For the latter two splits, we ensured no overlap sentences between the training and test sets for the held-out data. 

As shown in Tab.~\ref{tab:abl_splits}, the model maintained good performance when trained and tested on separate recording sessions, demonstrating its ability to generalize across sessions. NeuSpeech significantly outperformed random baselines on splits with no training-test overlap, validating its generalizability despite low text similarity between train and test sentences. Also we present sample outputs, revealing the model can still capture some semantic meaning and words when generalizing, though it struggles to produce fully correct novel sentences.


\section{Evaluation on words}



\label{supp.eval_words}
To figure out if our model is learning MEG signal, we designed this experiment. We randomly clip segments of varying lengths and use the corresponding words to train, and then we test on every single words. It turns out our model gets 5.8\% accuracy, since the test set has 2251 words, the chance level is 0.04\%, our model's word classification accuracy is 130 times the chance level. This means our model has learned something.

\section{Modification}
\label{supp.modification}
\begin{table}[h]
  \centering
  \caption{Performance of dropping the point-wise convolution layer. - indicates deletion of module.}
\label{tab:modification}
\resizebox{0.5\columnwidth}{!}{
\begin{tabular}{@{}llllllll@{}}
\toprule
\multicolumn{1}{c}{\multirow{2}{*}{Model}} & \multicolumn{4}{c}{BLEU-N (\%)} & \multicolumn{3}{l}{ROUGE-1 (\%)} \\ \cmidrule(l){2-8} 
\multicolumn{1}{c}{}                         & N=1    & N=2    & N=3   & N=4   & F         & P         & R        \\ \cmidrule(r){1-8}
baseline                                    & 41.65  & 36.47  & 33.51 & 31.39 & 44.86     & 46.71     & 44.74    \\
- conv                                   & 34.81  & 28.11  & 24.42 & 22.01 & 37.28     & 39.91     & 37.17    \\ \bottomrule
\end{tabular}
}
\end{table}

In line with \cite{D_fossez_2023_meg_eeg_clip_pretrain_meta_brain2speech}, we modified the whisper encoder by replacing the first convolution layer with two separate convolution layers to increase dimensionality and reduce temporal width. Comparing our base model with and without the point-wise convolution layer (Table~\ref{tab:modification}), we observed a significant drop in performance without the layer. This highlights the crucial role of the point-wise convolution layer in achieving good results by capturing important data features and representations.

The modification code is shown below.
\begin{lstlisting}[caption=Load whisper base model with its weights and modify the original model structure., label=supp.model_modification]
import torch.nn as nn
model = WhisperForConditionalGeneration.from_pretrained("openai/whisper-base")
d_model = kwargs['d_model']
conv1 = nn.Sequential(
    nn.Conv1d(kwargs['meg_ch'], d_model, kernel_size=3, padding=1),
    nn.GELU(),
    nn.Conv1d(d_model, d_model, kernel_size=3, stride=2, padding=1),
)
conv1.stride = (2,)
model.model.encoder.set_input_embeddings(conv1)
\end{lstlisting}

\section{More examples}
\label{supp.more_examples}
Here we show more generated samples on Gwilliams and Schoffelen dataset to show our model's capabilities. Also, we show the eeg-to-text model's performance here.

\newpage
\begin{lstlisting}[caption=NeuSpeech generation on GWilliams dataset~\cite{Gwilliams_2023_dataset_meg_208sensors_27persons_56h}. ,label=supp.neuspeech_generation_gwilliams]
start********************************
Predicted: Right
True: Right
end==================================

start********************************
Predicted:  The rental car records pointed to lacquered cherry-schedung and stood in the center of his ordered chamber in motel from Acres comb in?
True: Arthur replaced the tray onto the lacquered cherry table that stood in the center of his ordered chamber
end==================================

start********************************
Predicted: Why do you disregard my warnings my careful advice given in both of the best of friendship and designed only only enough to save you from sinking into this very-type debauchery?
True: Why do you disregard my warnings my careful advice given in the best of friendship and designed only to save you from sinking into this very type of corruption this literary debauchery
end==================================

start********************************
Predicted:  his hand blurred and snapped out as he stood back up
True: His hand blurred and snapped out as he stood back up
end==================================

start********************************
Predicted: Like I was saying
True: Like I was saying
end==================================

start********************************
Predicted:  He walked a wheel ruts hands in his pockets and shoes collecting dust
True: He walked the wheel ruts hands in his pockets and shoes collecting dust
end==================================

start********************************
Predicted: Tucker liked hit and liked see blood blood
True: Tucker liked hit and liked see blood
end==================================

start********************************
Predicted:  My patience was long gone and I was back in the car. But when I heard that many of you were looking for whatever it was, but what about this?
True: My patience was long gone and I was back in the car to warming up when Acres tapped on the window and told me he had found whatever he was looking for
end==================================

start********************************
Predicted: Arthur rose and shook his hand in
True: Arthur rose and shook his hand
end==================================

start********************************
Predicted: stone
True: stone
end==================================

\end{lstlisting}
\newpage
\begin{lstlisting}[caption=NeuSpeech generation samples on Schoffelen dataset~\cite{Schoffelen_2019_dataset_meg_273sensor_96person_80h}, label=supp.neuspeech_schoffelen_generations]
start********************************
Predicted: De fruitige dranken die soms vreemd smaakten werden geschonken op het feest.
True: De fruitige dranken die soms vreemd smaakten werden geschonken op het feest
end==================================

start********************************
Predicted: De magere dronkelap die de uitsmijter ontloopt heeft iets te verbergen.
True: De firma viert haar tienjarig bestaan met een gigantisch feest
end==================================

start********************************
Predicted: Het is een minder klimaat tot bijna halverwege de maand van houdt aan gereden
True: Het is een minder tijdrovende procedure dan klassieke animatie
end==================================

start********************************
Predicted: De vers gemaakte salade die deze bedorven tomaat verpest heeft gooi ik weg.
True: De vers gemaakte salade die deze bedorven tomaat verpest heeft gooi ik weg
end==================================

start********************************
Predicted: Het z'n ouders verperen in het poed dat vertragen gaten de moeile bloemen met een graag.
True: Zonder groot protest nam het meisje dat moest huilen een hap groente
end==================================

start********************************
Predicted: Vrolijk spelend op hun paarse trommels liepen de leerlingen naar school.
True: De reporter die de vermoeide ballerina interviewt wil veel weten
end==================================

start********************************
Predicted: Kreta dat prachtige stranden heeft is vanwege het massale toerisme veel minder aantrekkelijk de ene te.
True: Kreta dat prachtige stranden heeft is vanwege het massale toerisme veel minder aantrekkelijk
end==================================

start********************************
Predicted: De rector die de stoere puber strafte was erg onredelijk erg erg en teliopad
True: De rector die de stoere puber strafte was erg onredelijk
end==================================

start********************************
Predicted: De secretaresse geeft de advocaat die de mannelijke beambte adviseert.
True: in dat kamp voor tieners was geen enkele begeleider die verstand van zaken had
end==================================

start********************************
Predicted: De oude agent die het verdachte pakketje ontwalingig zorgs
True: Zelfs een grove dwaling zorgde niet voor een mentale inzinking
end==================================

\end{lstlisting}

\newpage

\begin{lstlisting}[caption=EEG-to-text model~\cite{wang2022open_aaai_eeg2text} generation without teacher forcing., label=supp.eeg_to_text_generation_wotf]
start################################################
Predicted: He said he had no idea how long it would take him to get back home
True: Right
end################################################
start################################################
Predicted: He said he had no idea how long it would take him to get there
True: Arthur replaced the tray onto the lacquered cherry table that stood in the center of his ordered chamber
end################################################
start################################################
Predicted: It was a long time since I had last seen him and he looked like an older version of the man who used to be my father
True: Why do you disregard my warnings my careful advice given in the best of friendship and designed only to save you from sinking into this very type of corruption this literary debauchery
end################################################
start################################################
Predicted: It was a long time since I had seen him in the flesh and he looked so much older than his twenty-five years
True: His hand blurred and snapped out as he stood back up
end################################################
start################################################
Predicted: It was a long time since I had last seen him and he looked like an old man in his late forties or early fifties
True: Like I was saying
end################################################
start################################################
Predicted: It was a long time since I had last seen him and he looked like an old man in his late forties or early fifties
True: He walked the wheel ruts hands in his pockets and shoes collecting dust
end################################################
start################################################
Predicted: It was a long time since I had been in the same room with anyone else and it felt good to be alone again
True: Tucker liked hit and liked see blood
end################################################
start################################################
Predicted: He said he had no idea how long it would take him to get back home
True: My patience was long gone and I was back in the car to warming up when Acres tapped on the window and told me he had found whatever he was looking for
end################################################
start################################################
Predicted: He said he had no idea how long it would take him to get back home
True: Arthur rose and shook his hand
end################################################
start################################################
Predicted: He said he had no idea how long it would take him to get back home
True: stone
end################################################
start################################################
Predicted: It was a long time since I had seen him in the flesh
True: Allan sat down at his desk and pulled the chair in close
end################################################
\end{lstlisting}

\newpage

\begin{lstlisting}[caption=EEG-to-text model~\cite{wang2022open_aaai_eeg2text} generation with teacher forcing. ,label=supp.eeg_to_text_generation_wtf]
start################################################
Predicted:  now... and and. and and and and to to to to to to to to to to to to to to to to to to to to to to to
True: Right
end################################################
start################################################
Predicted:  was the old with the tablekeyered table table and had in the middle of the room table
True: Arthur replaced the tray onto the lacquered cherry table that stood in the center of his ordered chamber
end################################################
start################################################
Predicted:  do you think the advice friend eyes the first of times my to to make my your a dark of timeauchery
True: Why do you disregard my warnings my careful advice given in the best of friendship and designed only to save you from sinking into this very type of corruption this literary debauchery
end################################################
start################################################
Predicted:  eyes was with he back of he reached up up
True: His hand blurred and snapped out as he stood back up
end################################################
start################################################
Predicted:  a said a and........
True: Like I was saying
end################################################
start################################################
Predicted:  was away length slowly and and his pockets looked in dust
True: He walked the wheel ruts hands in his pockets and shoes collecting dust
end################################################
start################################################
Predicted: ears was the the run hit
True: Tucker liked hit and liked see blood
end################################################
start################################################
Predicted:  father was running gone I was ready to the dark drive up Iacia me the window said me to was a a he was looking for
True: My patience was long gone and I was back in the car to warming up when Acres tapped on the window and told me he had found whatever he was looking for
end################################################
start################################################
Predicted:  was from walked his head
True: Arthur rose and shook his hand
end################################################
start################################################
Predicted: . was was was was was b b b.. to to to to to to to to to to
True: stone
end################################################

\end{lstlisting}

\newpage
\section{Ethics Statements}
\label{supp.ethics statements}
While promising for advancing AI, decoding thoughts from neurodata raises serious privacy, ethical and regulatory concerns if misapplied. We conduct this research openly and transparently, stressing that voluntary informed consent is vital for any future applications. Continued responsible development with strong safeguards is needed to ensure this capability is never abused or used to infringe on individuals' psychological autonomy, privacy or security without permission. As an emerging field at the interface of neuroscience and technology, clear guidelines and oversight also require discussion to adequately address the risks as this work progresses.

\section{Societal Impacts}
\label{supp.social_impact}
\textbf{Positive Societal Impacts}

\begin{itemize}
    \item \textbf{Enhanced Communication for Disabled Individuals:} The development of brain-to-text decoding technology holds significant promise for individuals with severe speech and motor impairments. By providing a non-invasive method for translating brain signals into text, this technology can vastly improve communication abilities, enabling these individuals to express their thoughts and needs more effectively and independently.
    
    \item \textbf{Advancements in Neurological Research:} This work contributes to a deeper understanding of the brain’s functioning, particularly how neural signals can be translated into language. Such advancements can fuel further research in neuroscience, potentially leading to new discoveries about brain processes and cognitive functions.
    
    \item \textbf{Innovations in Assistive Technology:} The insights and technologies developed from this research can drive the creation of new assistive devices and applications. These innovations could range from more sophisticated speech synthesis tools to advanced brain-computer interfaces that can improve the quality of life for people with disabilities.
    
    \item \textbf{Educational and Learning Tools:} Brain-to-text decoding can be leveraged to create new educational tools that adapt to individual cognitive states, enhancing learning experiences. This could lead to personalized education systems that cater to the specific needs and abilities of each student.
    
    \item \textbf{Increased Accessibility:} By making communication technology more accessible, this research can help bridge the gap for those who struggle with traditional communication methods. This can foster greater inclusion in various aspects of society, including education, employment, and social interactions.
\end{itemize}

\textbf{Negative Societal Impacts}

\begin{itemize}
    \item \textbf{Privacy and Security Concerns:} The ability to decode brain signals into text raises significant privacy issues. Unauthorized access to or misuse of this technology could lead to breaches of personal thoughts and intentions, posing serious ethical and legal challenges.
    
    \item \textbf{Potential for Misuse:} In the wrong hands, brain-to-text technology could be exploited for malicious purposes, such as surveillance, coercion, or manipulation. The potential for abuse highlights the need for strict regulatory frameworks and ethical guidelines.
    
    \item \textbf{Socioeconomic Disparities:} As with many advanced technologies, there is a risk that brain-to-text decoding could exacerbate existing socioeconomic disparities. Those with access to cutting-edge medical care and technology may benefit disproportionately, while underprivileged groups might be left behind.
    
    \item \textbf{Dependence on Technology:} Overreliance on brain-to-text technology could lead to decreased development and usage of natural communication skills in some individuals, particularly if the technology becomes widely adopted for everyday communication.
    
    \item \textbf{Ethical and Moral Dilemmas:} The ability to read and interpret brain signals brings up ethical questions about consent and autonomy. For instance, there could be situations where individuals are pressured to use such technology or where it is used without their explicit consent, raising serious moral concerns.
\end{itemize}

In conclusion, while the work on brain-to-text decoding presents remarkable opportunities for societal advancement, particularly in enhancing communication for individuals with disabilities, it also brings forth significant challenges and ethical considerations that need to be addressed. Balancing the benefits and risks will be crucial in ensuring that this technology is developed and implemented in a manner that maximizes its positive impacts while mitigating potential negative consequences.

\section{Safeguards}
\label{supp.safeguards}
\textbf{Data Privacy and Security}

To ensure the protection of individuals' brain data, several measures need to be implemented:

\begin{itemize}
    \item \textbf{Encryption:} All brain signal data should be encrypted both in transit and at rest to prevent unauthorized access. Advanced encryption standards should be utilized to safeguard the data.
    
    \item \textbf{Anonymization:} Before processing, data should be anonymized to remove any personally identifiable information. This reduces the risk of associating the data with specific individuals.
    
    \item \textbf{Access Control:} Strict access control mechanisms should be in place, ensuring that only authorized personnel can access the data. Multi-factor authentication and regular audits should be conducted to maintain security.
\end{itemize}

\textbf{Ethical Guidelines}

Establishing clear ethical guidelines is crucial to ensure responsible use of brain-to-text decoding technology:

\begin{itemize}
    \item \textbf{Informed Consent:} Participants must be fully informed about the nature of the research, how their data will be used, and the potential risks involved. Consent should be obtained explicitly and documented.
    
    \item \textbf{Usage Limitations:} Clear guidelines should outline permissible uses of the technology, prohibiting applications that could lead to coercion, manipulation, or unauthorized surveillance.
    
    \item \textbf{Ethical Review Boards:} All research and applications involving brain-to-text technology should be reviewed by ethical boards to ensure compliance with established ethical standards.
\end{itemize}

\textbf{Regulatory Compliance}

Compliance with legal and regulatory standards is essential to protect individuals and ensure ethical practices:

\begin{itemize}
    \item \textbf{Data Protection Laws:} Adherence to data protection laws, such as GDPR in Europe or HIPAA in the United States, is mandatory. These laws govern how data should be collected, stored, and processed.
    
    \item \textbf{Regular Audits:} Regular audits should be conducted to ensure compliance with legal and regulatory requirements. These audits can help identify and rectify any lapses in data handling practices.
    
    \item \textbf{Transparency Reports:} Organizations should publish transparency reports detailing how the technology is used, data protection measures in place, and any incidents of data breaches or misuse.
\end{itemize}

\textbf{Technological Safeguards}

Implementing technological safeguards can help prevent misuse and ensure the reliability of brain-to-text systems:

\begin{itemize}
    \item \textbf{Bias Mitigation:} Efforts should be made to detect and mitigate any biases in the data or the models. Diverse datasets and fairness checks can help in creating more equitable systems.
    
    \item \textbf{Robustness and Accuracy:} The models should be rigorously tested for accuracy and robustness. Continuous monitoring and updates can help maintain high performance and prevent errors.
    
    \item \textbf{Fail-Safe Mechanisms:} Implementing fail-safe mechanisms can prevent the system from generating potentially harmful or incorrect outputs. These mechanisms can halt operations if anomalies are detected.
\end{itemize}

\textbf{Public Engagement and Transparency}

Engaging with the public and maintaining transparency can build trust and ensure the technology is developed with societal needs in mind:

\begin{itemize}
    \item \textbf{Public Consultation:} Regular consultations with the public and stakeholders can provide valuable feedback and address concerns about the technology's use.
    
    \item \textbf{Educational Initiatives:} Informing the public about the technology, its benefits, and potential risks can help demystify the science and garner support for responsible development.
    
    \item \textbf{Open Communication:} Transparent communication about research findings, developments, and any incidents can help maintain public trust and ensure accountability.
\end{itemize}

In summary, implementing comprehensive safeguards encompassing data privacy, ethical guidelines, regulatory compliance, technological measures, and public engagement is vital to responsibly advance brain-to-text decoding technology. These safeguards will help mitigate risks, ensure ethical practices, and maximize the positive societal impacts of this innovative field.

\end{document}

%% file: text/table_score.tex
\begin{table*}[t]
\centering
\caption{\label{tb:scores} Evaluation metrics of MEG-to-Text translation under raw waves input. \textcolor{black}{ For a fair comparison, these results has \textit{w/} and \textit{w/o} the teacher-forcing evaluation setting since EEG-to-Text~\cite{wang2022open_aaai_eeg2text} only has teacher-forcing originally. Default is w/o teacher forcing and `\textit{w/ tf}' means with teacher-forcing.} }

\vspace{5pt}
\small
\resizebox{0.95\textwidth}{!}{
\centering
\begin{tabular}{@{}clllllllll@{}}
\toprule
\multicolumn{2}{c}{} & \multicolumn{4}{c}{\textbf{BLEU-N (\%) $\uparrow$}} & \multicolumn{3}{c}{\textbf{ROUGE-1 (\%)}$\uparrow$} & \multicolumn{1}{c}{\textbf{WER (\%) $\downarrow$}} \\
 \cmidrule(lr){3-6} \cmidrule(lr){7-9} \cmidrule(lr){10-10}
\textbf{Modality} & \textbf{Method} & N=1 & N=2 & N=3 & N=4 & F  & P & R & \\
\midrule
\rowcolor[HTML]{EFEFEF}
Speech & Whisper~\cite{radford2023robust_whisper_model_originalpaper} & 67.79 & 59.15 & 51.97 & 45.78 & 75.90 & 73.53 & 81.24 & 41.21 \\
Noise & NeuSpeech & 0.0 & 0.0 & 0.0 & 0.0 & 2.84 & 23.9 & 1.5 & 99.98 \\
MEG & eeg-to-text~\cite{wang2022open_aaai_eeg2text} & 9.21 & 2.13 & 0.57 & 0.14 & 9.74 & 10.73 & 11.38 & 118.25 \\
MEG & NeuSpeech & 57.68 & 52.76 & 48.84 & 45.42 & 60.24 & 61.86 & 61.92 & 58.89 \\ 
\midrule
MEG & eeg-to-text~\cite{wang2022open_aaai_eeg2text} \textit{w/ tf} & 21.66 & 9.27 & 3.93 & 1.73 & 20.11 & 22.48 & 18.71 & 95.69 \\
MEG & NeuSpeech \textit{w/ tf} & 78.13 & 71.25 & 64.79 & 58.86 & 81.91 & 80.52 & 83.68 & 25.38 \\
\bottomrule
\end{tabular}
}
\vspace{-15pt}
\end{table*}

%% file: text/table_targets.tex
\begin{table}[!t]
    \centering
    \caption{Transcription results on the unseen MEG waves, where \textbf{bold} denotes a correct match between ground truth and our prediction. \underline{Underline} denotes a fuzzy match with similar semantic meanings. 
    \textcolor{black}{
    For a fair comparison, these results \textbf{do not keep} the teacher-forcing evaluation setting as EEG-to-Text~\cite{wang2022open_aaai_eeg2text} and Dewave \cite{duan2023dewave_brain2text}. This means the decoding process can hold on accumulated errors and predicts the next token with model-generated tokens from previous steps.  } More samples are provided in supp.~\ref{supp.more_examples}.
    \label{tab:generation_results} }
    
    \resizebox{1\textwidth}{!}{
    \begin{tabular}{l }
         
     \toprule 
     
     \textbf{Decoding Results on \textit{GWilliams}~\cite{Gwilliams_2023_dataset_meg_208sensors_27persons_56h}} \\ \midrule
     
     Ground Truth: At best they tales at worst expeditions into macabre realms no healthy mind need ever see \\
     Prediction: \textbf{At best} it is macred \textbf{at worst expeditions into macabre realms no healthy mind need ever see} that \\\midrule
     
     Ground Truth: He turned to regard the pillar still rising behind his house and ran to see what had been done \\
     Prediction: \textbf{He}-\textbf{turned to regard a pillar still rising behind his house and ran} away from...\textbf{to see what had been done}. \\ \midrule\midrule
     \textbf{Decoding Results on \textit{Schoffelen}~\cite{Schoffelen_2019_dataset_meg_273sensor_96person_80h}} \\ \midrule
     
     Ground Truth: Tijdens de rumoerige bespreking beslisten de leden de staking voort te zetten \\
     Prediction: \textbf{Tijdens de rumoerige bespreking beslisten de leden} of \textbf{de staking voort}. \\\midrule
     
     Ground Truth: Op die goede prestatie was helemaal niets af te dingen \\
     Prediction: \textbf{Op die goede prestatie was} naar een varken afkomen van het nieuwe afguk...
     
     
    
    \\\bottomrule
    \end{tabular}
    }
\end{table}


%% file: text/table_targets_comparison.tex
\begin{table*}[!t]
    \small
    \centering
    \caption{
    Transcription results comparison with different methods. \textbf{Bold} font means exact match of words. \underline{Underline} for similar but not exact matching. w/ means with. tf means teacher-forcing, pt means pretraining, jt means joint-training. eeg-to-text~\cite{wang2022open_aaai_eeg2text} here is using beam search as the same setting of ours which is described in Sec.~\ref{subsec:implementation details}. More samples are provided in supp.~\ref{supp.more_examples}.
    }
    \label{tab:generation_results_comparison} 
    
    \resizebox{0.95\textwidth}{!}{
\begin{tabular}{@{}llp{11cm}@{}}
\toprule
\\ \cmidrule(l){1-3} 
\multicolumn{1}{c}{\multirow{7}{*}{(1)}} & GT                & I seen him since high school maybe twenty years before and we were never buddies in the first place                                                             \\ 
\multicolumn{1}{c}{}                     & NeuSpeech         & \textbf{I seen him since high school} \underline{when I was young}, at least \textbf{before and we were never buddies in} any place.                              \\
\multicolumn{1}{c}{}                     & NeuSpeech w/ tf   & \textbf{i seen him since high school} when even\textbf{twenty years} \textbf{before and we were never buddies in} one\textbf{the first place}                                           \\
\multicolumn{1}{c}{}                     & NeuSpeech w/ pt   &  \textbf{I seen him since high school maybe twenty years before and we were never buddies in the first place} he talked to me since t.                                                                           \\
\multicolumn{1}{c}{}                     & NeuSpeech w/ jt   & \textbf{I seen him since high school maybe twenty years before and we were never buddies in the first place}                                                                                \\
\multicolumn{1}{c}{}                     & eeg-to-text~\cite{wang2022open_aaai_eeg2text}       & \underline{It was a long time since I had last seen him} in the flesh                                                                                                       \\
\multicolumn{1}{c}{}                     & eeg-to-text~\cite{wang2022open_aaai_eeg2text} w/ tf & was the and I \textbf{school} a \textbf{years} ago that he had both close \textbf{the first place}                                                                               \\\cmidrule(l){1-3} 
\multirow{7}{*}{(2)}                     & GT                & My patience was long gone and I was back in the car to warming up when Acres tapped on the window and told me he had found whatever he was looking for               \\
& NeuSpeech         & \textbf{My patience was long gone and I was back in the car}. But when I heard that many of you were \textbf{looking for} whatever it was, but what about this?                        \\
& NeuSpeech w/ tf   & \textbf{My patience was long gone and I was back in} a\textbf{the car}.\textbf{to warming up when Acres} was\textbf{tapped on the}the \textbf{window and} opened\textbf{told me he had found whatever he was looking for}. \\
& NeuSpeech w/ pt   & \textbf{My patience was long gone and I was back in the car}.                                                                                                                 \\
& NeuSpeech w/ jt   & \textbf{My patience was long gone and I was back in the car}. The window was looking good when Acres is ongoing on her car and she now has...                                 \\
& eeg-to-text~\cite{wang2022open_aaai_eeg2text}       & \underline{He said} he had no idea how \textbf{long} it would take him to get \textbf{back} home                                                                                                   \\
& eeg-to-text~\cite{wang2022open_aaai_eeg2text} w/ tf & father was running gone I was ready to the dark \underline{drive} up Iacia me the \textbf{window} said me to was a a \textbf{he was looking for}                                                   \\ \cmidrule(l){1-3} 
\multirow{7}{*}{(3)}                     & GT                & Very well said Arthur then take your story to that and let him defend you when no civilized publisher will approach you                                              \\
& NeuSpeech         & \textbf{Very well said Arthur}! Tell me \textbf{your story}, and I'll \textbf{defend} it from \textbf{you when no civilized publisher will approach you}.                                                \\
& NeuSpeech w/ tf   & \textbf{Very well said Arthur}!\textbf{then} that\textbf{take your story to} and\textbf{that and let him defend you} and \textbf{no civilized publisher will approach you} when                                   \\
& NeuSpeech w/ pt   & \textbf{Very well said Arthur then take your story to that and let him defend you when no civilized publisher will approach you} \textbf{very well}                                    \\
& NeuSpeech w/ jt   & \textbf{Very well said}! It was that and we can make sure that it and not \textbf{let him defend you when no civilized publisher will approach you}                                    \\
& eeg-to-text~\cite{wang2022open_aaai_eeg2text}       & It was the first time I had seen him in a long time and it made me very nervous                                                                                      \\
& eeg-to-text~\cite{wang2022open_aaai_eeg2text} w/ tf & little then he the seat and the place go it take it he one man \underline{would you}   \\ \bottomrule                                                                                         
\end{tabular}
    }
\end{table*}

%% file: text/table_abl_dataset.tex
\begin{wraptable}{r}{0.5\textwidth}
\setlength{\tabcolsep}{3.9pt}
\caption{Different datasets. \textbf{Bold} for higher socres.}
\centering
\resizebox{1.0\linewidth}{!}{
\begin{tabular}{@{}lllllllll@{}}
\toprule
\multicolumn{1}{c}{\multirow{2}{*}{Dataset}} & \multicolumn{4}{c}{BLEU-N (\%) $\uparrow$} & \multicolumn{3}{c}{ROUGE-1 (\%) $\uparrow$}  \\ \cmidrule(lr){2-8}
\multicolumn{1}{c}{}                         & N=1    & N=2    & N=3   & N=4   & F         & P         & R        &                           \\ \midrule
\textit{GWilliams}                          & \textbf{60.3} & \textbf{55.26} & \textbf{51.24} & \textbf{47.78} & \textbf{58.73} & \textbf{60.88} & \textbf{59.76}                       \\
\textit{Schoffelen}                                   & 52.89  & 48.53    & 46.02  & 44.18  & 54.28     & 53.87     & 56.14          \\ \bottomrule

\end{tabular}
}
\vspace{-15pt}
\label{tab:abl_datasets}
\end{wraptable}

%% file: text/table_split_datasets.tex
\begin{table}[h]
\caption{Different dataset splits. Noise means input model with Gaussian noise with same length of MEG signal. We shows the BLEU-1 score and an example of generation. \underline{Underline} denotes a fuzzy match with similar semantic meanings. Note, for hold-out story and sentences, we used nucleus sampling p=0.25 for generation. }
\begin{tabular}{@{}lllp{9.5cm}@{}}
\toprule
Hold-out                   & Noise                & \multicolumn{2}{l}{Generation and example sentence}                                                                                                                                                                                                                                                             \\ \midrule
\multirow{2}{*}{Session}   & \multirow{2}{*}{0.0} & \multirow{2}{*}{53.16} &  
\textbf{Ground truth}: \underline{something} worth reading worth writing                                                                                                                                                                                                                   \\
                           &                      &                        & 
\textbf{Prediction}: \underline{some} kind of what i sell                                                                                                                                                                                                                                          \\\midrule
\multirow{2}{*}{Story}     & \multirow{2}{*}{0.0} & \multirow{2}{*}{6.91}  &  \textbf{Ground truth}: I looked down and saw he was pointing to another marker just slightly \underline{different from the one we had just left}                                                                                              \\
                           &                      &                        &  
\textbf{Prediction}: It was different from now and it just wasn't working well. \\ \midrule     
\multirow{2}{*}{Sentences} & \multirow{2}{*}{0.0} & \multirow{2}{*}{6.54}  & \textbf{Ground truth}: her first officer \underline{was staring at} his panel                                                                                                                                                                                                                        \\
                           &                      &                        &  
\textbf{Prediction}: he \underline{glared at} allan eyes him and demanding a response but he                                                                                                                                                                                                       \\ \bottomrule
\end{tabular}
\label{tab:abl_splits}
\end{table}

%% file: neurips_2024.bbl
\begin{thebibliography}{10}

\bibitem{Metzger_2023_ecog_hubert_birnn_brain2speech_brain2text_avatar}
Sean~L. Metzger, Kaylo~T. Littlejohn, Alexander~B. Silva, David~A. Moses, Margaret~P. Seaton, Ran Wang, Maximilian~E. Dougherty, Jessie~R. Liu, Peter Wu, Michael~A. Berger, Inga Zhuravleva, Adelyn Tu-Chan, Karunesh Ganguly, Gopala~K. Anumanchipalli, and Edward~F. Chang.
\newblock A high-performance neuroprosthesis for speech decoding and avatar control.
\newblock {\em Nature}, 620(7976):1037–1046, August 2023.

\bibitem{Willett_2023_ecog_speech_neuroprosthesis_rnn_brain2speech2text}
Francis~R. Willett, Erin~M. Kunz, Chaofei Fan, Donald~T. Avansino, Guy~H. Wilson, Eun~Young Choi, Foram Kamdar, Leigh~R. Hochberg, Shaul Druckmann, Krishna~V. Shenoy, and Jaimie~M. Henderson.
\newblock A high-performance speech neuroprosthesis.
\newblock January 2023.

\bibitem{wang2022open_aaai_eeg2text}
Zhenhailong Wang and Heng Ji.
\newblock Open vocabulary electroencephalography-to-text decoding and zero-shot sentiment classification.
\newblock In {\em Proceedings of the AAAI Conference on Artificial Intelligence}, volume~36, pages 5350--5358, 2022.

\bibitem{duan2023dewave_brain2text}
Yiqun Duan, Charles Zhou, Zhen Wang, Yu-Kai Wang, and Chin teng Lin.
\newblock Dewave: Discrete encoding of eeg waves for eeg to text translation.
\newblock In {\em Thirty-seventh Conference on Neural Information Processing Systems}, 2023.

\bibitem{dash2020decoding_meg_phrases}
Debadatta Dash, Paul Ferrari, and Jun Wang.
\newblock Decoding imagined and spoken phrases from non-invasive neural (meg) signals.
\newblock {\em Frontiers in neuroscience}, 14:290, 2020.

\bibitem{csaky2023interpretable_meg_many_class}
Richard Csaky, Mats~WJ van Es, Oiwi~Parker Jones, and Mark Woolrich.
\newblock Interpretable many-class decoding for meg.
\newblock {\em NeuroImage}, 282:120396, 2023.

\bibitem{ghazaryan2023trials_MEG_decode_written_text}
Gayane Ghazaryan, Marijn van Vliet, Aino Saranp{\"a}{\"a}, Lotta Lammi, Tiina Lindh-Knuutila, Annika Hult{\'e}n, Sasa Kivisaari, and Riitta Salmelin.
\newblock Trials and tribulations when attempting to decode semantic representations from meg responses to written text.
\newblock {\em Language, Cognition and Neuroscience}, pages 1--12, 2023.

\bibitem{D_fossez_2023_meg_eeg_clip_pretrain_meta_brain2speech}
Alexandre Défossez, Charlotte Caucheteux, Jérémy Rapin, Ori Kabeli, and Jean-Rémi King.
\newblock Decoding speech perception from non-invasive brain recordings.
\newblock {\em Nature Machine Intelligence}, 5(10):1097–1107, October 2023.

\bibitem{radford2023robust_whisper_model_originalpaper}
Alec Radford, Jong~Wook Kim, Tao Xu, Greg Brockman, Christine McLeavey, and Ilya Sutskever.
\newblock Robust speech recognition via large-scale weak supervision.
\newblock In {\em International Conference on Machine Learning}, pages 28492--28518. PMLR, 2023.

\bibitem{xi2023unicorn}
Nuwa Xi, Sendong Zhao, Haochun Wang, Chi Liu, Bing Qin, and Ting Liu.
\newblock Unicorn: Unified cognitive signal reconstruction bridging cognitive signals and human language.
\newblock {\em arXiv preprint arXiv:2307.05355}, 2023.

\bibitem{Gwilliams_2023_dataset_meg_208sensors_27persons_56h}
Laura Gwilliams, Graham Flick, Alec Marantz, Liina Pylkkänen, David Poeppel, and Jean-Rémi King.
\newblock Introducing meg-masc a high-quality magneto-encephalography dataset for evaluating natural speech processing.
\newblock {\em Scientific Data}, 10(1), December 2023.

\bibitem{Schoffelen_2019_dataset_meg_273sensor_96person_80h}
Jan-Mathijs Schoffelen, Robert Oostenveld, Nietzsche H.~L. Lam, Julia Uddén, Annika Hultén, and Peter Hagoort.
\newblock A 204-subject multimodal neuroimaging dataset to study language processing.
\newblock {\em Scientific Data}, 6(1), April 2019.

\bibitem{papineni2002bleu_orig}
Kishore Papineni, Salim Roukos, Todd Ward, and Wei-Jing Zhu.
\newblock Bleu: a method for automatic evaluation of machine translation.
\newblock In {\em Proceedings of the 40th annual meeting of the Association for Computational Linguistics}, pages 311--318, 2002.

\bibitem{lin2004rouge_orig}
Chin-Yew Lin.
\newblock Rouge: A package for automatic evaluation of summaries.
\newblock In {\em Text summarization branches out}, pages 74--81, 2004.

\bibitem{Tang_2022_fmri_gpt_beam_search_brain2text}
Jerry Tang, Amanda LeBel, Shailee Jain, and Alexander~G. Huth.
\newblock Semantic reconstruction of continuous language from non-invasive brain recordings.
\newblock September 2022.

\bibitem{zhang2023adaptive_adalora}
Qingru Zhang, Minshuo Chen, Alexander Bukharin, Pengcheng He, Yu~Cheng, Weizhu Chen, and Tuo Zhao.
\newblock Adaptive budget allocation for parameter-efficient fine-tuning.
\newblock {\em arXiv preprint arXiv:2303.10512}, 2023.

\bibitem{bai2023dreamdiffusion}
Yunpeng Bai, Xintao Wang, Yan-pei Cao, Yixiao Ge, Chun Yuan, and Ying Shan.
\newblock Dreamdiffusion: Generating high-quality images from brain eeg signals.
\newblock {\em arXiv preprint arXiv:2306.16934}, 2023.

\bibitem{he2022masked_mae}
Kaiming He, Xinlei Chen, Saining Xie, Yanghao Li, Piotr Doll{\'a}r, and Ross Girshick.
\newblock Masked autoencoders are scalable vision learners.
\newblock In {\em Proceedings of the IEEE/CVF conference on computer vision and pattern recognition}, pages 16000--16009, 2022.

\end{thebibliography}
